# Blind restoration for non-uniform aerial images using non-local Retinex model and shearlet-based higher-order regularization

Rui Chen,[a] Huizhu Jia,[a,*] Xiaodong Xie,[a] Wen Gao[a]

[a]Peking University, National Engineering Laboratory for Video Technology, No. 5, Yiheyuan Road, Beijing, China, 100871

**Abstract**. Aerial images are often degraded by space-varying motion blur and simultaneous uneven illumination. To recover high-quality aerial image from its non-uniform version, we propose a novel patch-wise restoration approach based on a key observation that the degree of blurring is inevitably affected by the illuminated conditions. A non-local Retinex model is developed to accurately estimate the reflectance component from the degraded aerial image. Thereafter the uneven illumination is corrected well. And then non-uniform coupled blurring in the enhanced reflectance image is alleviated and transformed towards uniform distribution, which will facilitate the subsequent deblurring. For constructing the multi-scale sparsified regularizer, the discrete shearlet transform is improved to better represent anisotropic image features in term of directional sensitivity and selectivity. In addition, a new adaptive variant of total generalized variation is proposed for the structure-preserving regularizer. These complementary regularizers are elegantly integrated into an objective function. The final deblurred image with uniform illumination can be extracted by applying the fast alternating direction scheme to solve the derived function. The experimental results demonstrate that our algorithm can not only remove both the space-varying illumination and motion blur in the aerial image effectively but also recover the abundant details of aerial scenes with top-level objective and subjective quality, and outperforms other state-of-the-art restoration methods.

*****Corresponding author:** Huizhu Jia**,** E-mail: hzjia@pku.edu.cn

## 1 Introduction

In current airborne long-distance imaging systems, aerial images can be digitally captured from large terrestrial areas by using CMOS or CCD sensors. However, these observed images often suffer non-uniform illumination and motion blurring, simultaneously. In the scenario of aerial imaging, the degree of blurring depends on multiple unpredictable factors such as uneven illumination, relative motion and camera shaking[1]. Specifically, due to both the random change of light transmission path and the limited irradiance collection of image sensors, the captured scenes with a large dynamic range will exhibit low contrast, dark or saturated regions, where the image intensities would be strongly affected in terms of the uneven distribution[2]. More importantly, the non-uniform illumination can complicate motion blur in the exposure duration. In addition, the poorer environmental lighting can result in the more severely blurring. If there is a relative motion between the camera and the scene during the exposure time, the space-varying blur will be generated in the resulting image. Similarly, the frequent vibrations of airborne platform cause the camera shaking, and would further blur the obtained images[3]. These non-uniform degradations of the aerial image can be unifiedly described in a mathematical model. Assuming that the airborne optical imaging system is spatially incoherent and linear[4], then the blurring process under varying lighting conditions is modeled as

$$\mathbf{g} = \mathbf{K}\hbar(\mathbf{f}_0) + \mathbf{n}. \quad (1)$$

where $\mathbf{g}$ denotes the given blurry and noisy image involving the uneven illumination. $\mathbf{f}_0$ denotes the latent clear and intrinsic reflectance image. The image noise $\mathbf{n}$ is modeled as the additive white Gaussian noise. The matrix $\mathbf{K}$ is a blur kernel or point spread function (PSF) of blur effect in an optical imaging system, where each sub-block matrix may correspond to a different low-pass filter for spatially-variant blurring process. The operator function $\hbar$ represents the illuminated modulation of other complex imaging factors. In practice, the blind restoration techniques are required to deblur input aerial image and enhance image quality to some extent. These techniques have wide applications in unmanned aerial vehicle (UAV) photography, ground surveillance, target detection and environmental exploration. The objective of blind restoration is to seek the best estimations of $\mathbf{f}_0$ and $\mathbf{K}$ from known degraded version $\mathbf{g}$. To solve this ill-posed inverse problem, certain prior information of the original image and PSF is used to regularize the recovery process by exploiting the intrinsic properties such as the smoothness, continuity, sparsity and probabilistic distributions for image structures.

The existing image restoration algorithms can be roughly categorized into three groups according to prior information about the blur kernel. With respect to the first group, the underlying blur kernel has already been known or accurately estimated through the response characteristics of an optical imaging system. Then this restoration problem can be reformulated as a non-blind image deconvolution (NBID) process. To produce a unique and stable restoration solution, a number of NBID algorithms have been proposed. Based on certain optimal estimation criterion, this problem can be solved by the classical Richardson-Lucy algorithm[5,6], Wiener filter[7] and the ordinary constrained least squares techniques[8] with low computation complexity. However, these methods often generate the low-quality restoration results containing the ringing artifacts at high noise levels. More advanced regularization strategies for the restoration tasks include the sparsity constraints[9], total variation (TV) and its variants[10-12], the wavelet and tight frame transforms[13,14], and the probabilistic models on image features[15]. The great efforts have been made to take full advantage of the sparsity in image spaces. Shao et al.[9] proposed to use the combination of the $\ell_0$ norm and the $\ell_2$ norm as sparsity measure which penalizes high-frequency components of natural scenes. The TV-based deconvolution method is able to find the approximate solutions to differential equations in bounded variation spaces. To overcome the undesired staircase artifacts caused by standard TV model, Hu et al.[11] derived two partial image derivatives as isotropic and anisotropic higher-order TV (HDTV) penalties to enhance line-like image features and preserve the singularities in the image. Based on structure tensor, Chierchia et al.[12] constructed the nonlocal total variation regularization to penalize nonlocal variations, which have the capability to capture first-order information and provide more robust measurement of image variations. Many multi-scale transforms for image representation and analysis are popular for image restoration based on analytical and synthesized operators. Cai et al.[14] established a framelet system to solve the deconvolution problem by extending the wavelet frame. The natural image containing random textures can be taken as a realization of the estimated probability distribution. Niknejad et al.[15] recovered the degraded images using a multivariate Gaussian mixture model (GMM) as a prior which is built upon the accumulation of similar patches in neighborhood. Currently, the sparse representation schemes by learning dictionaries from the example images have become popular in recovering the blurred image. Zhang et al.[16] established a group-based sparse representation framework and deblurred the image by enforcing the local sparsity and non-local self-similarity on the image patches. Papyan et al.[17] proposed a multi-scale restoration method which imposed the same patch-based Gaussian model on different scale patches extracted from the image and improved the deblurring performance.

In many practical situations, the PSF of an optical imaging system can not exactly be known due to the insufficient information of measurement. Therefore, the restoration methods in the second group have been developed to estimate both the blur kernel and the latent image from an input blurred image. In general, the filtering property of blur kernel is assumed to be space-invariant in this blind image deconvolution (BID) problem. The successes of the BID methods arise from imposing the reasonable prior knowledge on the PSF including positiveness, known shape, smoothness, symmetry, or known finite support. Under the predictable lighting conditions, the PSF of an imaging platform can be easily obtained by measuring the response of knife-edge regions in the degraded image with low to middle precision[18]. To achieve wider applications and higher accuracy, most blind restoration methods focus on the mathematically construction of the optimization functional. Fergus et al.[19] integrated a mixture Gaussian model on the gradient magnitudes into a variational Bayesian framework and then the uniform motion-blur was removed by minimizing the cost function. Xu et al.[20] adopted shock filter to adaptively predict the salient edge map for kernel initialization and then restored motion blurred images via iterative refinement scheme. The frequency spectrum of the PSF can be utilized to recover the final kernel. Goldstein et al.[21] developed a power-law model together with an accurate spectral whitening formula to estimate the power spectrum of the blur and then recovered the PSF based on the modified phase retrieval algorithm. Liu et al.[22] proposed a convex kernel regularizer by exploring the spectrum change as a convolution operator and then estimated the desired kernel by minimizing this regularizer. Recently, the Radon transform of the spectrum of the blurred image has been proposed for motion blur estimation. This idea is that the angle and length of linear uniform motion are estimated from the image cepstral features in the Radon space[23]. To improve the final deblurring result, multiple images have been jointly applied for providing the additional information. Yuan et al.[24] obtained the accurate kernel by combining information extracted from a pair of images with the complementary exposure time. Zhang et al.[25] presented a coupled penalty function to adapt the quality of multiple observed images, and estimated the latent sharp image and blur kernel by optimizing this function.

Unfortunately, the assumption on spatially-invariant PSF would not be satisfied in many imaging systems, and this can cause large restoration errors in BID method. Hence, the restoration methods in the third group have been further developed when the blur kernel varies across the image plane. Due to the 3D rotation of the camera, the motion blur is significantly non-uniform across the image. The typical analysis[26] for the spatially-varying motion blur is to segment a blurred image into several uniform regions, and the restored results are obtained with the combination of uniform blur kernels. Whyte et al.[27] presented a parameterized geometric model of the blurring process in term of the motion orientation, and substituted this model into the existing deblurring algorithms. Zhang et al.[28] modeled the blur kernel as a series of parameterized projective transform matrices and estimated the latent clear image by incorporating the projection models. To achieve an efficient non-uniform deblurring algorithm, Yu et al.[29] identified the erroneous PSFs by measuring the similarity between the neighboring kernels, and employed the total variation regularization to recover the latent sharp image after replacing the erroneous kernel. Recently, some learning-based works have been proposed for blur removal. Sun et al.[30] predicted the Markov random field model of motion blur by learning a convolutional neural network and removed the non-uniform motion blur with this patch-level model. If the variations on scene irradiance have the wide dynamic range, the need for longer exposure time increases the possibility of motion-blur degradation. Vijay et al.[31] used the transformation spread function to model the motion blur and handled non-uniform blur by minimizing the cost functional derived from the blur kernel. As another important case, the fluctuations of incident light can strongly affect long-distance imaging systems, and further cause the space-varying blurs. This problem is generally reduced to a shift invariant one. To mitigate the varying effect, one restoration route[32] first employs an image registration technique to align images and then a deblurring process is applied to the combined image. The other route[2] detects lucky regions to suppress the geometric deformations by a local sharpness metric and then fuses them to produce a large high-quality deblurring image.

It is notable that none of the non-uniform deblurring algorithms have fully considerations on both the influence of uneven illumination and image characteristics mainly originating from real aerial scenes[33,34]. Actually, the restoration issue for non-uniform aerial images still remains unsolved. As for the deeper understanding, when the incident light comes from the ground surface, the uneven illumination often occurs in the aerial image because the different light sources with varying intensities arrive at the imaging sensors. Moreover, the non-uniform refraction index in the air layers can cause light beams through unexpected paths, and so the images taken through turbulence atmosphere would generate the distortions and blurring visually. Eventually, these complex factors result in the spatially non-uniform changes of image intensities which are greatly distinct from those of usual images. The captured landscape images are composed of diverse natural and man-made structures which exhibit wide range of image features including low-level edges, midlevel edge junctions, high-level object parts, and complete shapes. Although the compensation mechanisms of the aerial imaging system can mitigate the vibrations to some extent, the images captured by cameras attached to airplanes are still possible to be blurred by both the forward motion and vibrations. Therefore, the restoration for these features requires the more suitable representation and reconstruction schemes. At present, the Retinex theory has received particular attention and thus been widely developed to remove the uneven illumination and enhance the contrast. Most Retinex-based algorithms[35-39] extract the reflectance component as the enhanced result by isolating the illumination, and therefore they can enhance the details obviously. However, the uneven nature of the land surface often adversely affects the stability of the results. Since the various Retinex assumptions depend on the applications, we explore its non-local variant to correct the non-uniform illumination of the aerial image based on the knowledge about lighting variations. The traditional wavelet transform is not effective at dealing with singularities in the image. To overcome the limitations of wavelet, a new geometric analysis tool called shearlet transform[40-43] has been evolved and it can capture multi-scale and multi-directional line singularities of the image. In addition, it can preserve more edges and textures compared with other transforms. All of these properties make the shearlet transform an attractive candidate for image representation. To further improve the analysis for the complex geometric features in the aerial image, we extend the original discrete shearlet transform while retaining shift invariance and anisotropy. Higher-order regularization has become increasingly popular for tackling image restoration problem in recent years. Total Generalized Variation (TGV)[45-48], especially its second-order variant, has shown promising results as a robust regularization term. However, TGV suffers from the major drawback that it only penalizes the pixel values at the fixed directions. Moreover, purely TGV-based models are not able to accurately locate structural discontinuities. Hence, the new developments for TGV are mainly accounted for preserving various image structures. Based on above aspects, additionally because previous restoration methods for the aerial image may fail to remove both the space-varying illumination and motion blur under coupled degradations, our research would bridge the gap between the restored quality and limited prior information.

In this paper, we present a powerful blind restoration method to recover the non-uniformly motion blurred aerial image under uneven illumination. As shown in Fig. 1, the whole restoration procedure mainly includes three cascading stages. A new non-local Retinex model is developed to dramatically decrease the non-uniform illumination by the advantage of non-local similarity of image patches. At first stage, the estimated reflectance image will tend to have uniform illumination and exhibit higher contrast by iteratively solving the constructed variational function. Moreover, because uneven lighting conditions are related with the spatially-varying blurring formation, the non-uniform distribution of the coupled motion blur can be partly mitigated in the resulting reflectance image. We improve the directional selectivity and sensitivity of original discrete nonseparable shearlet transform (DNST) by using the fan filters with arbitrary frequency partitioning, which are more effective in analyzing the piecewise smooth images with rich geometric information. Moreover, a structure-adaptive total generalized variation (SA-TGV) is designed in order to describe the intensity variations of the smooth region more precisely. Due to the fact that each blur kernel corresponding to the patch image can be reasonably taken as the spatially invariance on aerial imaging conditions, the blurry reflectance image is further performed the patch-wise recovery. At second stage, by integrating the improved DNST and two-order SA-TGV as the regularizers of a devised optimization functional, the desired solutions for clear patch image and blur kernel can be found by using a fast iterative numerical procedure. Finally, the deblurred aerial image with corrected illumination is obtained by stitching all the restored patch images. In our framework, the informative sharp edges and fine texture details can be recovered well in challenging aerial scenarios.

The main contributions of this paper can be summarized as follows:

1. Based on the sparsity and fidelity priors, a non-local Retinex model is proposed to correct both uneven illumination and spatially-varying distribution of motion blur in the aerial image. Moreover, numerical optimization implementation for this model is given.

2. Through the selection of a particular image-driven basis and the nonseparable digitized realization of compactly supported framework, an improved DNST is constructed to provide accurate localization and sparsely encode anisotropic singularities of the aerial image.

3. In order to better preserve the geometric structures and differentiate with the noise, the SA-TGA regularization is developed and efficiently incorporated into an object function. Then a fast alternating direction scheme is adopted to solve this optimization function.

The paper is organized as follows: Section 2 describes the non-local Retinex model and the corresponding numerical implementation. In Section 3, the improved DNST is introduced and the novel SA-TGV is given in detail. The proposed restoration framework is presented in Section 4. The experimental results are illustrated in Section 5, and finally we conclude this work and discuss future research in Section 6.

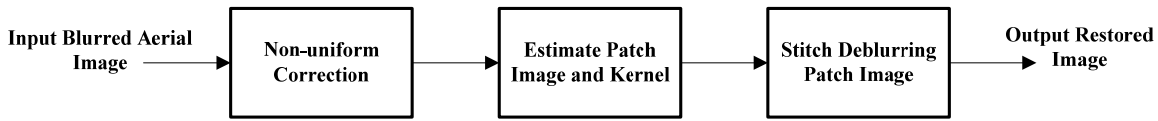

**Fig. 1** Block diagram of the proposed blind restoration method.

## 2   Non-uniform Illumination Correction

*2.1 Non-local Retinex Model*

In complex natural environment, the elevation of sun, the anisotropic irradiance of land surface and the random fluctuations of atmospheric turbulence could mutually affect the light conditions across the location. Eventually, because of the varying light attenuated and refraction from the solar spray, the acquired aerial images will incur the unavoidable non-uniform illumination in many cases. According to previous experimental results[32,33], the illumination power received by the camera conforms to the exponential decay of light power distant from the object. Different from smooth spatial variations in usual photography scenes, the environmental illumination distributions of a natural scene might contain abrupt spatial variations of local illumination as the result of their complex spatial structures, which produce shading, mutual reflection and occlusion[1]. Essentially, the physical model of aerial imaging formation can be seen as the process modulated by the environmental illumination. Based on this fact, the received aerial image at the sensors could be corrected by estimating reflectance component and removing the illumination effect. Currently, this process is elegant to be dealt with Retinex theory. The primary goal is to decompose the given image $\mathbf{g}$ into the product of the reflectance image $r$ and the illumination image $l$, which is given by

$$\mathbf{g} = r \cdot l. \qquad (2)$$

where $r$ depends on the physical characteristics of the material and is invariant to the illuminated and imaging conditions. $l$ is typically related with natural illumination distributions. Since these components represent different physical elements, this decomposition can isolate the desired image. In order to simplify the computation, the product form in Eq. (2) is converted to the logarithmic domain. Here, we define that $\tilde{\mathbf{g}}$, $\tilde{r}$ and $\tilde{l}$ are equal to $\log \mathbf{g}$, $\log r$ and $\log l$, respectively. Then taking the logarithm at both sides of this equation, it yields

$$\tilde{\mathbf{g}} = \tilde{r} + \tilde{l}. \tag{3}$$

According to this decomposition model, removing the uneven illumination from $\tilde{\mathbf{g}}$ is known to be a mathematically ill-posed problem. To guarantee the existence and uniqueness of the solution, prior assumptions are required to regularize both $\tilde{r}$ and $\tilde{l}$ simultaneously. The outdoor illumination $\tilde{l}$ in the large area can be considered as spatially smoothness. Furthermore, the reflectance image $\tilde{r}$ is exploited at known limited range, and should hold on sharp edges and textural details, which are perceptually pleasing. In addition, the average intensity of homogeneous area can be perceived as gray world assumption. Based on the observation that a reflectance image can be described by a number of local texture structures, non-adjacent pixels within same texture are assumed to have similar reflectance intensities. The non-local constraint on surface reflectance is introduced by measuring the similarity of two pixels. Furthermore, this constraint throughout the image relates the reflectance values of distant pixels, which favors the global consistency to the intrinsic decomposition. Then we utilize the coherence of the neighboring pixels to enforce the same behavior which can strengthen the regularization within the same regions and reduce its influence across the edges. In order to obtain solutions of higher accuracy, non-local formulation of total variation is employed to preserve abundant textures and sharp edges while preventing oversmoothness and blocky effect. Finally, we incorporate all these deduced constraints directly into an object function to formulate our non-local Retinex model as follows:

$$U(\tilde{r}) = \left\| \nabla \tilde{r} - \nabla \tilde{\mathbf{g}} \right\|_2^2 + \eta_0 \sum_{\mathbf{x} \in \Omega} \left[ \tilde{r}(\mathbf{x}) - \ln 0.5 \right]^2 + \eta_1 \sum_{\mathbf{x} \in \Omega} \sqrt{\sum_{\mathbf{y} \in G_\mathbf{x}} w(\mathbf{x},\mathbf{y}) \left[ \tilde{r}(\mathbf{x}) - \tilde{r}(\mathbf{y}) \right]^2}. \tag{4}$$

where $\Omega \subset \mathbb{R}^2$ is defined as the image domain. $(\mathbf{x}, \mathbf{y})$ denotes different pixel locations in the image domain. $\eta_0$ and $\eta_1$ are the positive real parameters for balancing the contribution of different constraint terms. The symbol $\nabla$ denotes the gradient operator. $G_\mathbf{x}$ is a selected neighboring region at pixel $\mathbf{x}$ of size $L \times L$. In practice, the relatively sharp region is selected for adapting the strength of uneven illumination through local sharpness metrics. The image intensity variance $\hat{v}$ is used for selecting the region $G_{\mathbf{x}}$ as follows:

$$\hat{v} = \frac{1}{L^2 - 1} \sum_{G_{\mathbf{x}}} \left[ r(\mathbf{x}) - \bar{r} \right]^2. \tag{5}$$

where $\bar{r}$ represents the mean value of image intensities in region $G_{\mathbf{x}}$. The sharp region is selected by maximizing the metric $\hat{v}$ within the certain size range. The function $w(\mathbf{x}, \mathbf{y})$ denotes the weight which is used to measure the mutual information between two image patches.

The weight $w(\mathbf{x}, \mathbf{y})$ is computed not only based on the relative distance of the two pixels but also based on the non-local similarity of their intensity values between two image patches in the reflectance $\tilde{r}$. Then this weight function is given by:

$$w(\mathbf{x}, \mathbf{y}) = \exp\left( -\frac{d(\mathbf{x},\ \mathbf{y})^2}{2h^2} \right). \tag{6}$$

where $d(\mathbf{x},\ \mathbf{y})$ is the Euclidean distance between pixels $\mathbf{x}$ and $\mathbf{y}$. The standard deviation $h$ acts as a filtering parameter and its magnitude controls the influence of intensity similarity and spatial proximity. The final image $r$ is combined with isoplanatic regions containing space-invariant illumination.

*2.2 Numerical Implementation for Correction*

In this section, we have presented the numerical implementation algorithm for solving the minimization of objective function in Eq. (4), which can generate the corrected result for uneven illumination. Considering physical characteristics of reflectance in the spatial domain, each value in the image $r$ is normalized to $0 < r \leq 1$. Thus, as equivalent constraint, the inequality $\tilde{r} \leq 0$ is also added. Then the proposed Retinex model is reformulated as the following constrained optimization problem:

$$\min_{\tilde{r}} U(\tilde{r}), \quad s.t.\ \tilde{r} \leq 0. \tag{7}$$

The closed-form solution for $\tilde{r}$ can be found with the standard optimization method. The desired solution is to guarantee the whole sequence convergence to the minimizer of the original problem. In addition, the constraints should hold automatically in the iteration process, and meanwhile, the illumination is reconstructed efficiently. For notational simplicity, we occasionally represent $\tilde{r}(\mathbf{x})$, $\tilde{r}(\mathbf{y})$ and $\tilde{g}(\mathbf{x})$ as $\tilde{r}_\mathbf{x}$, $\tilde{r}_\mathbf{y}$ and $\tilde{g}_\mathbf{x}$, respectively. Here, the classic steepest decent method is adopted to get the Euler-Lagrange equation as follows:

$$\begin{cases} \dfrac{\partial U}{\partial \tilde{r}_\mathbf{x}} = 2\Delta(-\tilde{r}_\mathbf{x} + \tilde{g}_\mathbf{x}) + 2\eta_0(\tilde{r}_\mathbf{x} - \ln 0.5) + \eta_1 \dfrac{\partial V}{\partial \tilde{r}_\mathbf{x}} = 0, \\ V(\tilde{r}_\mathbf{x}) = \sqrt{\sum_{\mathbf{y} \in G_\mathbf{x}} w(\mathbf{x},\mathbf{y})(\tilde{r}_\mathbf{x} - \tilde{r}_\mathbf{y})^2}. \end{cases} \quad (8)$$

where $\Delta$ is the Laplacian operator, which can be performed by a linear convolution with the kernel $[0\ 1\ 0; 1\ -4\ 1; 0\ 1\ 0]$. This operator is a Gaussian smoothing operation with increasing variance of the initial condition. Based on the Eq. (8), with respect to the constant iteration step, the gradient descent flow equation involving k-th iteration can be obtained and further discretized as

$$\begin{cases} \tilde{r}_\mathbf{x}^{k+1} = \tilde{r}_\mathbf{x}^k + \tau\left[ 2\Delta(-\tilde{r}_\mathbf{x}^k + \tilde{g}_\mathbf{x}) + 2\eta_0(\tilde{r}_\mathbf{x}^k - \ln 0.5) + \eta_1 \sum_{\mathbf{x} \in \Omega} w(\mathbf{x},\mathbf{y})(\tilde{r}_\mathbf{y}^k - \tilde{r}_\mathbf{x}^k)\omega_\mathbf{x}^k \right], \\ \omega_\mathbf{x}^k = \dfrac{\left|\nabla_w \tilde{r}_\mathbf{x}^k\right| + \left|\nabla \tilde{r}_\mathbf{y}^k\right|}{\left|\nabla_w \tilde{r}_\mathbf{x}^k\right|\left|\nabla \tilde{r}_\mathbf{y}^k\right|}, \quad \nabla_w \tilde{r}_\mathbf{x}^k = \sqrt{\sum_{\mathbf{y} \in G_\mathbf{x}} w(\mathbf{x},\mathbf{y})(\tilde{r}_\mathbf{y}^k - \tilde{r}_\mathbf{x}^k)^2}. \end{cases} \quad (9)$$

where $\tau$ is the iteration step length between iteration $\tilde{r}_\mathbf{x}^{k+1}$ and $\tilde{r}_\mathbf{x}^k$. When doing the implementation, the initialization value of Eq. (9) is set as the input image and the weight $w(\mathbf{x},\mathbf{y})$ is estimated from the resulting image at the previous iteration. The iteration will not stop until the terminal condition is satisfied. Through analyzing the iteration number, this process can quickly converge to a stable solution within the maximum iteration number $iter_{max}$. In iterative process, the uniform illumination is reconstructed and the structural details are preserved efficiently.

## 3    Regularization Strategy

*3.1 Improvement for DNST*

Compared with capability of the wavelet transform in describing the geometric structures and the distributed discontinuities of a multi-dimensional signal, the shearlet transform is a multi-scale and multi-directional representation system that can provide more geometrical information and optimally sparse approximation[42]. The 2-D continuous transform of a signal $f \in L^2(\mathbb{R}^2)$ is defined as

$$SH_\psi(f)(a,s,t) = \langle f, \psi_{a,s,t} \rangle; \quad \psi_{a,s,t}(x) = |\det M_{as}|^{-\frac{1}{2}} \psi(M_{as}^{-1} x - t). \tag{10}$$

where the shearlet system $\{a \in \mathbb{R}^+, s \in \mathbb{R}, t \in \mathbb{R}^2, x = (x_1, x_2) \in \mathbb{R}^2\}$ is generated by the operations of dilation, shear transformation, and translation of function. Let $B_s$ be a shear operator and $A_a$ be an anisotropic dilation matrix. Here each matrix $M_{as}$ is a product of $B_s$ and $A_a$. Then its formulation is given by

$$M_{as} = \begin{bmatrix} a & \sqrt{a}s \\ 0 & \sqrt{a} \end{bmatrix} = \begin{bmatrix} 1 & s \\ 0 & 1 \end{bmatrix} \begin{bmatrix} a & 0 \\ 0 & \sqrt{a} \end{bmatrix} := B_s A_a. \tag{11}$$

The generating function $\psi \in L^2(\mathbb{R}^2)$ can be appropriately chosen from compactly supported wavelets when satisfying the sufficient admissibility conditions to generate a frame[42]. So the shearlet systems form a tight frame of well-localized waveforms at various scales, directions and locations controlled by $a$, $s$ and $t$, respectively. Due to their good analytic and geometrical properties, the continuous shearlet with invariant directionality can accurately capture the anisotropic features in the image. In fact, the shearlet coefficients with large magnitudes are associated with spatially singularities such as edges, and the decay parameters across scales can be used to distinguish different image structures.

Through sampling the continuous shearlet transform (10) on appropriate discretizations of the scaling $a$, shear $s$ and translation $t$, the corresponding parameters $j$, $k$ and $m$ can be obtained for the discrete shearlet associated to a Parseval frame. To do this, we parameterize the matrices $A_a$ by dyadic numbers and $B_s$ by integers and replace the continuous translation variable by a point in the discrete lattice $\mathbb{Z}^2$.
Choosing $a = 2^{-j}$ and $s = -k$ from Eq. (10), we obtain the discrete shearlet system

$$\{\psi_{j,k,m} = |\det A_0|^{j/2} \psi(B_0^k A_0^j \cdot -m) : j,k \in \mathbb{Z}, m \in \mathbb{Z}^2\}. \tag{12}$$

where we let $a = 4$ and $s = 1$ for more specific matrices in Eq. (11). Then $A_0 = \begin{bmatrix} 4 & 0 \\ 0 & 2 \end{bmatrix}$ and $B_0 = \begin{bmatrix} 1 & 1 \\ 0 & 1 \end{bmatrix}$ provide directional windows in the space and frequency domain, which can be elongated along arbitrary directions. In particular, the appropriate choices of shearlet generators $\psi_{j,k,m}$ will guarantee stable reconstruction from the shearlet coefficients. In frequency domain, this regular discrete shearlet system can provide a nonuniform angular covering of the frequency plane by applying scaling and shear matrices when restricted to the finite discrete setting for implementation.

One can construct compactly supported shearlet frames by separable generating function in Eq. (12). Compared with the utilization of separable functions, the shearlets generated from nonseparable functions can more effectively cover the frequency plane and provide the better frame bounds. To improve the discrete shearlet transform in term of the selectivity and sensitivity, the nonsubsampled directional filter bank (NDFB) is designed to allow arbitrary directional frequency partitioning for the wedge-shaped subbands according to the intrinsic geometric characteristics of images. The frequency domain is divided into the basically vertical (BV) and basically horizontal (BH) subsets. When applying the 2D discrete pseudopolar Fourier transform (PPFT)[44] in accordance with the distribution of NDFB, the frequency parts are partitioned into several nonuniform rectangle subbands. Then we design two separable filters $P_{BV}(m_x, m_y)$ and $P_{BH}(m_x, m_y)$ computed from a pyramid filter $Q(k_1, k_2)$. The results are expressed as

$$\begin{cases} P_{BV}(m_x, m_y) = \sum_{k_1=0}^{N-1} \sum_{k_2=0}^{N-1} Q(k_1, k_2) \exp\left[-i(k_1 \frac{2\pi m_x m_y}{N^2} + k_2 \frac{\pi m_y}{N})\right] \\ P_{BH}(m_x, m_y) = \sum_{k_1=0}^{N-1} \sum_{k_2=0}^{N-1} Q(k_1, k_2) \exp\left[-i(k_1 \frac{\pi m_y}{N} + k_2 \frac{2\pi m_x m_y}{N^2})\right] \end{cases} \tag{13}$$

where two subsets BV and BH are defined on $\{-N \leq m_x \leq N, -N/2 \leq m_y \leq N/2, N \in \mathbb{Z}^+\}$. $Q(k_1, k_2)$ is half-band filter with diamond support. The fan-shaped response can be obtained from a diamond-shaped response by simple modulation in the frequency domain.

After decomposing the filters $P_{BV}(m_x, m_y)$ and $P_{BH}(m_x, m_y)$ into several subbands with rectangle supports, each support corresponds to a wedge-shaped region in the Cartesian frequency system. To further extract the directional information, the filters in Eq. (13) should be combined and filtered together. The combination $P(\xi_1, \xi_2)$ of two filter banks can be achieved in the following form:

$$P(\xi_1, \xi_2) = \begin{cases} P_{BV}(\xi_1 + \frac{3N}{2}, \xi_2), & -2N \leq \xi_1 \leq -N, -N \leq \xi_2 \leq N \\ P_{BH}(-\frac{N}{2} - \xi_1, \xi_2), & -N \leq \xi_1 \leq 0, -N \leq \xi_2 \leq N \end{cases} \quad (14)$$

where the designed fan filters $P_{BV}(m_x, m_y)$ and $P_{BH}(m_x, m_y)$ should be further transformed into the Cartesian Fourier formulation in coordinate $(\xi_1, \xi_2)$ and hold the period of $2\pi$ along the horizontal coordinate in order to divide the combination along the slope direction. This transformation results have arbitrary frequency partitioning and can extract the directional frequency distributions of images.

The basic shearlet generator is not a good choice for enhancing the directional sensitivity in the image because the fixed wedge-shaped subbands can not match the contours and textures in arbitrary directions. The proposed nonseparable fan filter in Eq. (14) can ensure highly directional sensitivity and good spatial localization, and its wedge shaped support is well adapted for covering the frequency domain. To improve the directional selectivity, the nonseparable shearlet generator $\psi_{j,k,m}^{non}$ is adopted to provide better frame bounds as well as better directional selectivity. Finally, the discrete $P(\xi_1, \xi_2)$ can be applied to construct the nonseparable shearlet filter $\psi_{j,k,m}^{non}$ by satisfying the generating condition for DNST[43].

$$\left\{\psi_{j,k,m}^{non}(\xi) = P(\xi_1, \xi_2)\psi_{j,k,m}(\xi) : \xi = (\xi_1, \xi_2)\right\}. \quad (15)$$

The improved DNST is constructed by from a separable compactly supported shearlet generator and the 2D nonseparable fan filter. In order to tailor to specific task for optimally representing features in nonuniform aerial images, the formulation of DNST at $j$-th subband is given as follows

$$DNST_j(f^d)(n) = (f^d * \psi_{j,k,m}^{non})(2^{J-j}c_1^j n_1, 2^{J-j/2}c_2^j n_2). \tag{16}$$

where $f^d \in \ell^2(\mathbb{Z}^2)$ is the sampled discrete data from the signal $f$ in the continuum domain. $\{c_1, c_2 \in \mathbb{R}^+\}$ are sampling constants for translation. $\{n = (n_1, n_2) \in \mathbb{Z}^2\}$ is an integer lattice point. Eq. (16) can map the signals or images $f^d$ to the sequence of shearlet coefficients for $j = 0, \ldots, J-1$. Based on the fast Fourier transform without the additional computational cost, $j$-th subband of the DNST can be efficiently computed in frequency domain by component-wise multiplication[43].

*3.2 Second-Order SA-TGV*

By incorporating smoothness from the partial derivatives of various orders, the TGV regularization generalizes TV and leads to piecewise polynomial intensities. It is capable of preserving sharp edges without the staircase-like effects of the bounded variation functional[45]. Moreover, the gradual intensity transitions in smooth regions are well preserved, and the piecewise affine function can be reconstructed in high-order derivative space. Because the numerical experiments show that the third or higher order TGV does not improve image quality enough to be worth the extra computing cost[46], the second-order $TGV^2$ regularizer is used to provide a good tradeoff between computational complexity and reconstruction accuracy without loss of good properties. For a vectorized form of image $\mathbf{f} \in \mathbb{R}^N$, the discretized formulation of $TGV^2$ can be written as

$$TGV^2(\mathbf{f}) = \min_{p \in BV(\Omega)} \sum_{\mathbf{x}=1}^{N} \alpha_1(\mathbf{x}) \|\nabla \mathbf{f}(\mathbf{x}) - p(\mathbf{x})\|_1 + \sum_{\mathbf{x}=1}^{N} \alpha_0(\mathbf{x}) \|\varepsilon(p)(\mathbf{x})\|_1. \tag{17}$$

where $\alpha_0(\mathbf{x})$ and $\alpha_1(\mathbf{x})$ are positive weights varying in every pixel $\mathbf{x}$. $p$ is the vector representation of symmetric tensor fields in bounded variation space $BV(\Omega)$. To efficiently solve the Eq. (17), the directional operators $\nabla \mathbf{f}$ and $\varepsilon(p)$ can be further approximated by

$$\nabla = [D_1, D_2]^\mathrm{T}, \ \varepsilon(p) = \begin{bmatrix} D_1 p_1 & \frac{1}{2}(D_2 p_1 + D_1 p_2) \\ \frac{1}{2}(D_2 p_1 + D_1 p_2) & D_2 p_2 \end{bmatrix}. \tag{18}$$

where $D_1$ and $D_2$ are the circulant matrices corresponding to the forward finite difference operators with periodic boundary conditions along the vertical and horizontal directions, respectively. And $p = [p_1, \ p_2]^\mathrm{T}$ is an approximation of first-order gradient $\nabla \mathbf{f}$ in the image $\mathbf{f}$. This reformulation of TGV makes its convexity property computationally feasible. For all pixels $\mathbf{x} \in \{1, 2, \ldots, N\}$, $\alpha_0(\mathbf{x})$ and $\alpha_1(\mathbf{x})$ can be set according to the different derivative characteristics in the image.

Although the effectiveness of TGV model has been demonstrated by favoring the piecewise affine solutions, its performance suffers from the major drawback that it is still sensitive to the weighting setting of parameters. In the original TGV, the identical weight is often used to penalize second-order derivatives of all the pixels in the image. However, aerial images are composed of inhomogeneous components and these components possess different derivative characteristics from each other. Therefore, the identical setting for weights of all pixels will lead to edge blurring and structural loss in the restored results. The second-order SA-TGV as a non-trivial extension is designed to remedy these problems by adapting the weighting parameters which locally depend on the locations and directions of image structures. We employ the structure tensor to extract information about the local geometry of image in a neighborhood of each pixel. The smoothed structure tensor $T_s$ is given by

$$T_s(\mathbf{f}) = \begin{bmatrix} G_\sigma * \mathbf{f}_1^2 & G_\sigma * \mathbf{f}_1 \mathbf{f}_2 \\ G_\sigma * \mathbf{f}_2 \mathbf{f}_1 & G_\sigma * \mathbf{f}_2^2 \end{bmatrix}, \ G_\sigma(\mathbf{x}) = \frac{1}{\sigma\sqrt{2\pi}} \exp(-\frac{|\mathbf{x}|^2}{2\sigma^2}). \tag{19}$$

where $T_s(\mathbf{f})$ is obtained from image $\mathbf{f}$ by convolution with a Gaussian kernel $G_\sigma(\mathbf{x})$ with variance $\sigma^2$. $\mathbf{f}_1$ and $\mathbf{f}_2$ are the directional derivative in the directions, respectively. Let the eigenvalues of $T_s(\mathbf{f})$ be $(\lambda_+(\mathbf{x}),\ \lambda_-(\mathbf{x}))$, with the corresponding eigenvectors $(v_+(\mathbf{x}),\ v_-(\mathbf{x}))$. The eigenvalues are ordered with the maximum and minimum $(|\lambda_+(\mathbf{x})| \geq |\lambda_-(\mathbf{x})|)$ respectively, which indicate the presence of image structures and average contrast within a neighborhood along the eigen-directions. Moreover, the eigenvectors $(v_+(\mathbf{x}),\ v_-(\mathbf{x}))$ can provide local directions which maximize the intensity fluctuations.

Based on structure tensor computed from Eq. (19), an indicator function $E(\mathbf{x})$ is considered to represent the anisotropic features of geometry and characterize different image structures. In particular, the magnitude relationship of two eigenvalues is adopted to indicate the fine structures such as edges, flat and corner regions. This function could measure and response to spatially varying structures in the image which contains high spatial frequencies. And then the salient structures will be present when the values of $E(\mathbf{x})$ are close to the decision threshold. By automatically self-tune the weighting strengths reflecting local image structures, we formulate the following indicator function,

$$E(\mathbf{x}) = \frac{\lambda_+(\mathbf{x},\ \sigma) - \lambda_-(\mathbf{x},\ \sigma)}{\chi + \lambda_+(\mathbf{x},\ \sigma) + \lambda_-(\mathbf{x},\ \sigma)}. \tag{20}$$

where $\chi > 0$ is a small free parameter for solution stability and its value is set in the range $\chi \in [0.005,\ 0.05]$. The parameter $\sigma > 0$ associated with the structural scales and noise levels has the effect of the pre-smoothing, which is important for getting the desired responses from computing derivatives. The eigenvalues $\lambda_+$ and $\lambda_-$ are computed at a specific scale for retaining and enhancing multiscale image structures. For computing $E(\mathbf{x})$ from Eq. (20), we first normalize the eigenvalues to the range $[0,\ 1]$.

If both eigenvalues $0 \ll \lambda_- \leq \lambda_+$ are large, it indicates the presence of the corner structures. In addition, an edge structure exists when the eigenvalues accord with $0 \approx \lambda_- \leq \lambda_+$. In the homogeneous or noisy regions, two eigenvalues satisfy the condition $\lambda_-, \lambda_+ \approx 0$. Therefore, based on the indicating function $E(\mathbf{x})$ responding to various local structures, we can accurately describe the adaptivity of the weighting parameters. By measuring the influence for informative structures, the weights are given by

$$\begin{cases} \alpha_0(\mathbf{x}) = E(\mathbf{x})\underline{\alpha}_0 + (1 - E(\mathbf{x}))\overline{\alpha}_0 \\ \alpha_1(\mathbf{x}) = E(\mathbf{x})\underline{\alpha}_1 + (1 - E(\mathbf{x}))\overline{\alpha}_1 \end{cases} \quad (21)$$

where the function $E(\mathbf{x})$ is limited to the range [0, 1]. The small parameters $\underline{\alpha}_0, \underline{\alpha}_1 > 0$ are used to control the relative ratio of two regularizations, which can determine the reconstruction quality to a great extent. Usually their values are chosen independent from the image content or noise level. The values of $\overline{\alpha}_0$ and $\overline{\alpha}_1 > 0$ depends mainly on the noise level of the image. Based on the adaptive weights derived from Eq. (21), the proposed SA-TGA can be achieved and represented by Eq. (17).

## 4 Proposed Restoration Method

In our restoration framework, the observed image **g** is first partitioned into fully overlapping patches $\{\mathbf{g}^{b_i}\}_{i=1}^n$. Then each patch $\mathbf{g}^b$ is corrected using the non-local Retinex model in Eq. (4), and the resulting image $r^b$ will tend to be the uniform distribution of illumination and space-invariant blur. At the same time, the image contrast has indeed been enhanced to facilitate the subsequent deblurring. After obtaining each uniform patch separately, the core step is to restore them by optimizing the derived reconstruction model. This model is constructed by incorporating SA-TGV regularizers in Eq. (17) and the improved DNST in Eq.(16). The numerical solutions can be obtained by using the alternating direction method of multipliers (ADMM)[49]. Finally, the clear and uniform patches are merged as the final aerial image by a plain averaging[2].

### 4.1 Restoration Formulation

The structured adaptive regularizers are used to jointly estimate the sharp image $\mathbf{f}_0$ and the blur kernel $\mathbf{K}$. In terms of the illumination and blur correction, the $\ell_1$-norm regularization is imposed on both the latent image $\mathbf{f}_0$ and the kernel $\mathbf{K}$ as the efficient sparsified constraints. In addition, the PSF of an aerial imaging system is always positive, upper bounded by the peak value of the diffraction-limited PSF, and normalized. For the reliable reconstruction of the latent image $\mathbf{f}_0$, the regularization scheme integrates both the SA-TGV regularizer and the DNST for multi-scale sparsity by full advantage of preserving directional features and high-order smoothness. Due to the sparsity-promoting features and adaption to different local structures, the coefficients of DNST are adopted as analysis-based prior terms. As a result, we propose the following reconstruction functional for solving ill-posed restoration problem in Eq. (1)

$$\min_{\mathbf{K},\mathbf{f}_0} \frac{1}{2}\|\mathbf{K}\hbar(\mathbf{f}_0)-\mathbf{g}\|_2^2 + \gamma\|\mathbf{K}\|_1 + \lambda\sum_{j=1}^{J}\|DNST_j(\mathbf{f}_0)\|_1 + \sum_{\mathbf{x}=1}^{N}\alpha_1(\mathbf{x})\|\nabla\mathbf{f}_0(\mathbf{x})-p(\mathbf{x})\|_1 + \sum_{\mathbf{x}=1}^{N}\alpha_0(\mathbf{x})\|\varepsilon(p)(\mathbf{x})\|_1. \qquad (22)$$

where $\gamma$, $\lambda$, $\alpha_0$ and $\alpha_1$ are the positive parameters. $\gamma$ can control smoothness of the PSF. $\lambda$ is a balancing factor relying on the gradients and the sparsity of the latent image. $DNST_j(\cdot)$ is $j$-th subband of the DNST by using fast numerical computation method[43]. $N$ denotes the total number of image pixels. $J$ is the total number of decomposed subbands.

To reduce the computation complexity, all the patches can be restored using the same model in parallel. The size of each patch $\mathbf{g}^b$ is selected according to the non-uniform illumination level of input aerial image. The corresponding kernel $\mathbf{K}^b$ is further transformed into the block-circulant matrix for performing the convolution operator. Moreover, the directional derivatives for original patch image $\nabla\mathbf{f}_0^b$ are approximated by $D\mathbf{f}_0^b$, where $D$ is the circulant matrix corresponding to the forward finite difference operators. The convolution operator for the improved DNST is simplified as the generating operator $\Psi$. At the deblurring stage, we consider that the corrected image $r^b$ is directly taken as the input to deduce the Eq. (22). Then the restoration model specified for image patch is rewritten as

$$\min_{\mathbf{K}^b, \mathbf{f}_0^b} \frac{1}{2}\|\mathbf{K}^b \mathbf{f}_0^b - \mathbf{r}^b\|_2^2 + \gamma \|\mathbf{K}^b\|_1 + \lambda \sum_{j=1}^{J^b} \|\Psi_j \mathbf{f}_0^b\|_1 + \sum_{\mathbf{x}=1}^{N^b} \alpha_1(\mathbf{x}) \|D\mathbf{f}_0^b(\mathbf{x}) - p(\mathbf{x})\|_1 + \sum_{\mathbf{x}=1}^{N^b} \alpha_0(\mathbf{x}) \|\varepsilon(p)(\mathbf{x})\|_1. \tag{23}$$

where $J^b$ is the decomposition number of all subbands, and $N^b$ is the total number of pixels in patch image. The adaptive weights $\alpha_0(\mathbf{x})$ and $\alpha_1(\mathbf{x})$ for each patch are set identical values.

## 4.2 Optimization Solutions Using ADMM

To solve the minimization optimization model in Eq. (23), we use the augmented Lagrangian and apply the ADMM. This method performs an iterative procedure that splits the constrained optimization into a series of convex functions, and then the sub-problems are solved by the fast iterative shrinkage-thresholding algorithm and a faster alternating minimization algorithm[49]. The corresponding constrained optimization problem of Eq. (23) is converted by adding the auxiliary variables and the quadratic penalty functions for each $\ell_1$-term. Then the equivalently constrained formulation of Eq. (23) is given as

$$\begin{cases} \min_{\mathbf{K}^b, \mathbf{f}_0^b, p, W_j, Y, V} \frac{1}{2}\|\mathbf{K}^b \mathbf{f}_0^b - \mathbf{r}^b\|_2^2 + \gamma \|\mathbf{K}^b\|_1 + \lambda \sum_{j=1}^{J^b} \|W_j\|_1 + \sum_{\mathbf{x}=1}^{N^b} \alpha_1(\mathbf{x}) \|Y(\mathbf{x})\|_1 + \sum_{\mathbf{x}=1}^{N^b} \alpha_0(\mathbf{x}) \|V(\mathbf{x})\|_1 \\ s.t. \quad W_j = \Psi_j \mathbf{f}_0^b, \quad Y(\mathbf{x}) = D\mathbf{f}_0^b(\mathbf{x}) - p(\mathbf{x}), \quad V(\mathbf{x}) = \varepsilon(p)(\mathbf{x}) \end{cases} \tag{24}$$

Note that the auxiliary variables $W_j$, $Y$ and $V$ are introduced without crossed other variables. Attaching the Lagrangian terms to the constraints, the simplified augmented Lagrangian function for Eq. (24) can be derived by the following form

$$\mathcal{L}(W_j, Y, V, \mathbf{K}^b, \mathbf{f}_0^b, p, \widetilde{W}_j, \widetilde{Y}, \widetilde{V}) = \begin{bmatrix} \frac{1}{2}\|\mathbf{K}^b \mathbf{f}_0^b - \mathbf{r}^b\|_2^2 + \gamma \|\mathbf{K}^b\|_1 + \lambda \sum_{j=1}^{J^b} \|W_j\|_1 + \sum_{\mathbf{x}=1}^{N^b} \alpha_1(\mathbf{x}) \|Y(\mathbf{x})\|_1 + \sum_{\mathbf{x}=1}^{N^b} \alpha_0(\mathbf{x}) \|V(\mathbf{x})\|_1 \\ + \frac{\beta_0}{2} \sum_{j=1}^{J^b} \|W_j - \Psi_j \mathbf{f}_0^b - \widetilde{W}_j\|_2^2 + \frac{\beta_1}{2} \|Y - (D\mathbf{f}_0^b - p) - \widetilde{Y}\|_2^2 + \frac{\beta_2}{2} \|V - \varepsilon(p) - \widetilde{V}\|_2^2 \end{bmatrix}. \tag{25}$$

where $\beta_0$, $\beta_1$, $\beta_2 > 0$ are the penalty parameters. Theoretically any positive values of these parameters ensure the convergence of ADMM, and the specific choice is used in the experiments. $\widetilde{W}_j$, $\widetilde{Y}$ and $\widetilde{V}$ are the scaled Lagrangian multipliers.

For fixed value of one variable, the updates of other variables are independent of one another. The ADMM minimizes Eq. (25) separately leading to subproblems which have closed-form solutions. The main subproblems can be grouped into three blocks. Using the shrinkage operator, three subproblems are similar and perform the following updates:

$$\begin{cases} W_j^{n+1} = \arg\min_{W_j} \lambda \|W_j\|_1 + \frac{\beta_0}{2} \|W_j - \Psi_j \mathbf{f}_0^b - \widetilde{W}_j\|_2^2 \\ Y^{n+1} = \arg\min_{Y} \sum_{\mathbf{x}=1}^{J^b} \alpha_1(\mathbf{x}) \|Y(\mathbf{x})\|_1 + \frac{\beta_1}{2} \|Y - (D\mathbf{f}_0^b|^n - p^n) - \widetilde{Y}^n\|_2^2 \\ V^{n+1} = \arg\min_{V} \sum_{\mathbf{x}=1}^{J^b} \alpha_0(\mathbf{x}) \|V(\mathbf{x})\|_1 + \frac{\beta_2}{2} \|V - \varepsilon(p^n) - \widetilde{V}^n\|_2^2 \end{cases} \quad (26)$$

The minimization with respect to $\mathbf{f}_0^b$ and $p$ is performed, thus $(\mathbf{f}_0^b, p)$-subproblem can be solved separately and sequentially. It leads to the following iteration:

$$\{\mathbf{f}_0^b|^{n+1}, p^{n+1}\} = \arg\min_{\mathbf{f}_0^b, p} \left[ \begin{array}{l} \frac{1}{2}\|\mathbf{K}^b|^n \mathbf{f}_0^b - \mathbf{r}^b\|_2^2 + \frac{\beta_0}{2} \sum_{j=1}^{J^b} \|W_j^{n+1} - \Psi_j \mathbf{f}_0^b - \widetilde{W}_j^n\|_2^2 \\ + \frac{\beta_1}{2} \|Y^{n+1} - (D\mathbf{f}_0^b - p) - \widetilde{Y}^n\|_2^2 + \frac{\beta_2}{2} \|V^{n+1} - \varepsilon(p) - \widetilde{V}^n\|_2^2 \end{array} \right]. \quad (27)$$

Based on the fact that the matrix $\mathbf{K}^b$ can be diagonalized under the Fourier transform, the optimal solution is obtained. The corresponding subproblem is of the form

$$\mathbf{K}^b|^{n+1} = \arg\min_{\mathbf{K}^b} \frac{1}{2}\|\mathbf{K}^b \mathbf{f}_0^b|^{n+1} - \mathbf{r}^b\|_2^2 + +\gamma \|\mathbf{K}^b\|_1. \quad (28)$$

To solve three subproblems for the Lagrangian multipliers, the updates can be done by using the Newton iterative operator. Thus the updates of these multipliers are expressed by

$$\begin{cases} \widetilde{W}_j^{n+1} = \widetilde{W}_j^n + \beta(\Psi_j \mathbf{f}_0^b|^{n+1} - W_j^{n+1}) \\ \widetilde{Y}^{n+1} = \widetilde{Y}^n + \beta(D\mathbf{f}_0^b|^{n+1} - p^{n+1} - Y^{n+1}) \\ \widetilde{V}^{n+1} = \widetilde{V}^n + \beta(\varepsilon(p^{n+1}) - V^{n+1}) \end{cases} \quad (29)$$

The convergence of numerical solutions for our restoration model depends on the classic ADMM because this problem is convex. The iteration conditions for linear convergence are guaranteed when $0 < \beta < (\sqrt{5}+1)/2$ is satisfied. The stopping criterion holds for the primal residual $\|res\|_2^2 \leq T_r$ and the tolerance $\left\| \mathbf{f}_0^b \right|^{n+1} - \mathbf{f}_0^b \right|^n \right\|_2^2 \Big/ \left\| \mathbf{f}_0^b \right|^n \right\|_2^2 \leq T_t$. The overall procedure of proposed restoration algorithm is summarized in Algorithm 1.

**Algorithm 1** The proposed blind restoration algorithm.

---
1: **Input**: The aerial image $\mathbf{g}$ with non-uniform illumination and motion blur, the noise variance $\sigma^2$.
2: **Initialization: adaptively select** the patch size and the parameters $(\overline{\alpha}_0, \underline{\alpha}_0, \overline{\alpha}_1, \underline{\alpha}_1)$, **set** all other parameters.
3: **Transform** the image $\mathbf{g}$ into the patches $\{\mathbf{g}^{b_i}\}_{i=1}^n$.
4: **While** a maximum iteration number $iter_{\max}$ is not satisfied **do**
5:   Estimate the reflectance image by solving Eq. (7)
6: **End While**
7: **Output**: the corrected image $\mathbf{r}^{b_i}$.
8: **Repeat:**
9:   Estimate the latent patch $\mathbf{f}_0^{b_i}$ by Eq. (27).
10:  Estimate the blur kernel $\mathbf{K}^{b_i}$ by Eq. (28).
11: **Until** meets the stopping criterion
12: **Merge** all the patches $\{\mathbf{f}_0^{b_i}\}_{i=1}^n$.
13: **Return** the restored image $\mathbf{f}_0$.
---

## 5 Experimental Results

### 5.1 Settings and Implementation Details

In this section, a series of testing experiments have been arranged and implemented to illustrate the effectiveness and efficiency of the proposed blind restoration framework, which is applied to handle the aerial image with both the non-uniform illumination and the motion blur. The whole image restoration tasks include the illumination correction, deblurring and denoising at the same time. To fulfill the more complete evaluation quantitatively, the test images in our experiments are selected from the synthetic data and real aerial image sets, in which the image sources involve the benchmark images commonly used as ground truth and aerial images captured by various airborne platforms. All the blurred images are assumed to be contaminated by additive Gaussian white noise with the zero mean. The objective quality of the restored image is evaluated using two assessment standards, namely the peak signal-to-noise ratio (PSNR) and the structural similarity index (SSIM). All algorithms are carried out using MATLAB 2010 with the Intel Core i7 CPU @ 2.4 GHz and 8 GB memory.

In the proposed framework, all important parameters need be set and initialized. In our practice, these involved parameters are determined by using two strategies. One is to tune manually some possible values and then set it to the optimal values with satisfactory restoration performance; the other is to apply the adaptive adjustment rules with an arbitrary initial guess. The parameters $\eta_0$ and $\eta_1$ are set as 0.01 and 0.02, respectively, where overall good results are found. The parameter $h$ is fixed to 2, which gives a good trade-off between accuracy and computational complexity. The initial size $L \times L$ of image patch is set as 40×40. The iteration length $\tau$ is selected at 0.08, which can achieve the good performance on the most test images. In the iteration process for non-uniform correction, $iter_{max}$ is fixed to 100 according to the convergence performance. For implementation of the DNST, we use the 17×17 maximally fan filters with eight directional subbands in every level to compute digital shearlet filters, which have four level decompositions across scales. It has been experimentally found that the value $\chi = 0.025$ is a reasonable choice. For each input image, the approximate optimal parameters ($\bar{\alpha}_0$, $\underline{\alpha}_0$, $\bar{\alpha}_1$, $\underline{\alpha}_1$) are determined by maximizing the PSNR value from the range of $10^{-3}$ to $10^{-2}$. The parameter $\gamma$ controls the sparsity of the kernel and is fixed to 0.25. The parameter $\lambda$ is related to the noise level, and is set to be 0.1 for the noisy data and 0.01 for all the noise-free data. Through the convergence analysis in section 4.2, we set $\beta_0 = 300$, $\beta_1 = 10^{-3}$, $\beta_2 = 10^{-5}$, $\beta = 1$ for all the numerical results. The initial size of the discrete PSF is set to $3 \times 3$ and the final size is chosen no more than $21 \times 21$. The initial estimate for $\mathbf{K}^b$ is set as the unit pulse. The initial solution for $\mathbf{f}_0^b$ is chosen as the pre-processed image $r^b$. Two thresholds are used to terminate the reconstruction iteration. The residual threshold $T_r$ is set as $10^{-4}$. The tolerance threshold $T_t$ is set as $10^{-5}$.

*5.2 Effectiveness Evaluation for Non-uniform Correction*

The non-uniform correction results are fully evaluated through three groups of experiments. The first group is to only testify the non-uniform correction capability for the varying illumination existing in the image. The second group is designed to clarify the relationship between the varying illumination and motion blur, and indicate the effectiveness for alleviating the blur. The third group is further to disclose the fact that uneven lighting conditions can couple and cause the more complex blur. Moreover, the correction performance for the non-uniform blur is measured.

The experiment on synthetic data was firstly carried out. To make convincing assessments to the correction results, three competitive methods are used to compare. An aerial image acquired from Washington DC is chosen as the original image shown in Fig. 2(a). It is horizontally degraded, as the test data shows in Fig. 2(b). The correction results of the mask[35], variational Retinex[37] (VR), TV-Retinex[36] (TVR) and our result are shown in Fig. 2(c)-(f), respectively. Obviously, Fig. 2(f) shows the best visual effect of all four methods because non-local TV constraint can favor to restore the more clear edges and details, and meanwhile, eliminate the blocky artifacts. However, other typical methods have limitations on rendering the image content without fully considering spatial similarity. In summary, the proposed method can efficiently remove the influence of uneven light and enhance the image details.

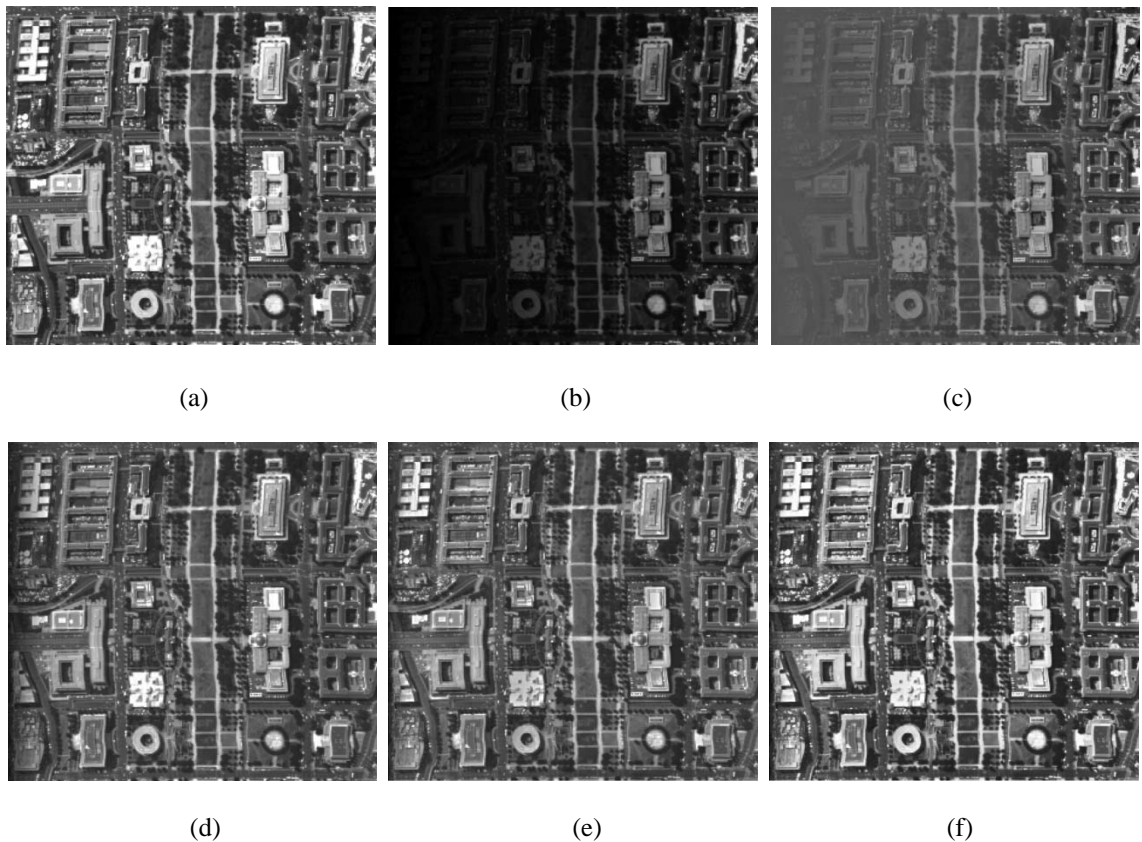

(a)                  (b)                  (c)

(d)                  (e)                  (f)

**Fig. 2** Varying illumination correction results on the synthetic data: (a) original aerial image, (b) degraded image, (c) mask filtering, (d) variational Retinex, (e) TV-Retinex, and (f) our result.

The second experiment was executed to compare the different deblurred results. In the nearly noise-free case, Fig. 3(b) and 3(e) are respectively obtained by convolving input images Fig. 3(a) and 3(d) by a synthetic 1-D delta filter with direction 10 degree and length 10 pixels. At deblurring step, a ground-truth approach[50] is employed to restore the sharp edges and details while simultaneously suppressing the image noise. For fair comparison, once the kernel has been estimated, the same parameters are used to reconstruct the final sharp image. As we can see from Fig. 3 of a college, the result in Fig. 3(f) is much better than that of Fig. 3(c), which produces too sharpen image and high contrast and reduces boundary ringing artifacts. This is mainly because the uneven illumination in the image can decrease the estimation accuracy of the blur kernel. In essence, it is concluded that relatively good quality image can be more reliably deblurred.

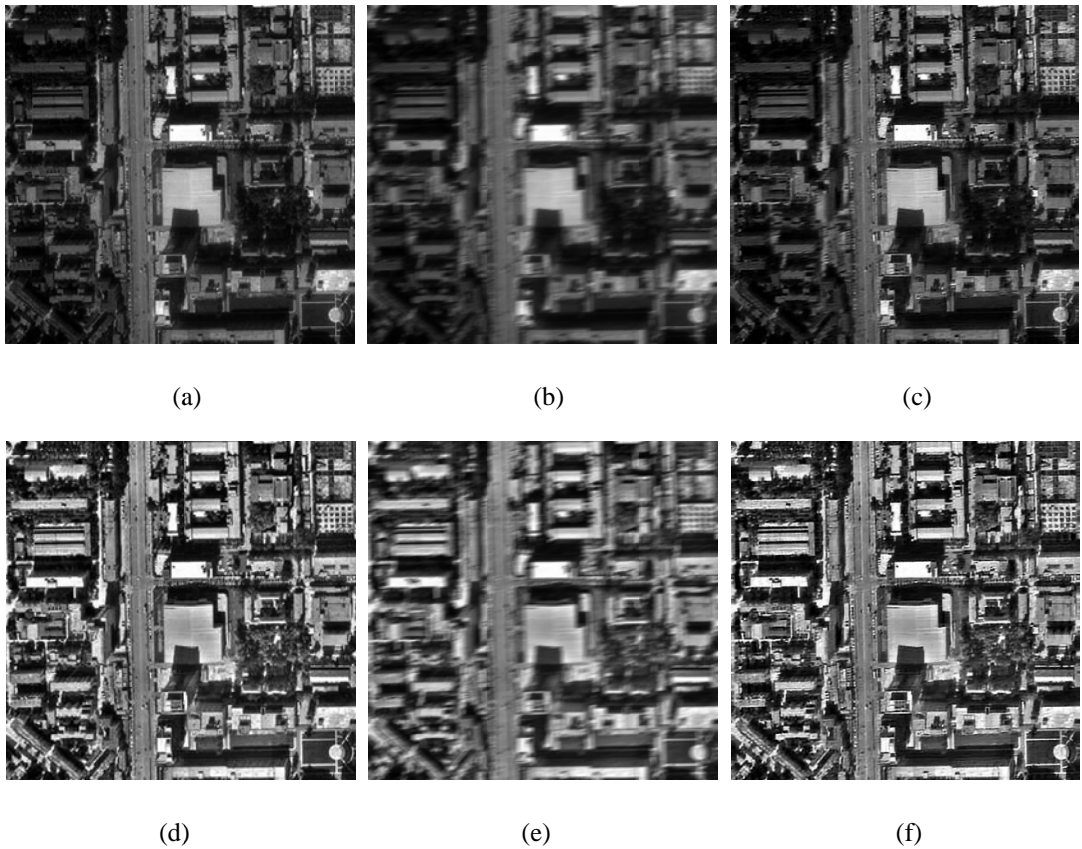

(a)          (b)          (c)

(d)          (e)          (f)

**Fig. 3** Deblurred comparison on the synthetic data: (a) original aerial image, (b) blurred image in (a), (c) deblurred result for (b), (d) corrected result for (a), (e) blurred image in (d), and (f) deblurred result for (e).

The third experiment was performed to further explore coupled relationship between the illuminated distributions and blurring formation. And the corrected performance for non-uniform motion blur is evaluated. Fig. 4(a) of a building was captured by placing the camera on our UAV, which can be manually controlled to achieve the complex motion. By analyzing the observation sequence, the noise in the obtained image can be modeled using the Gaussian distribution. It is reasonable to assume that motion blur will occur beyond the exposure setting. The image can be deteriorated by space-varying illumination. This coupled imaging process and non-uniform blur have been included into the real image Fig. 4(a). Note that Fig. 4(d) has been corrected for uniformly illumination throughout the scene. After employing the same deblurring approach in the second experiment, it can be seen from that the geometric structures in Fig. 4(e) are sharper than those of Fig. 4(b) while the deblurring result of Fig. 4(e) has less halo artifacts than that of Fig. 4(b). Because the corrected image can be recovered sharper edges, the PSFs are estimated more accurate from them. As shown in Fig. 4, the estimated kernels with different formulations in Fig. 4(c) have reflected the non-uniform blur. Because the kernels in Fig. 4(f) have almost same distributions, it is asserted that our correction model is valid for transforming to the nearly space-invariant blurs with a certain degree.

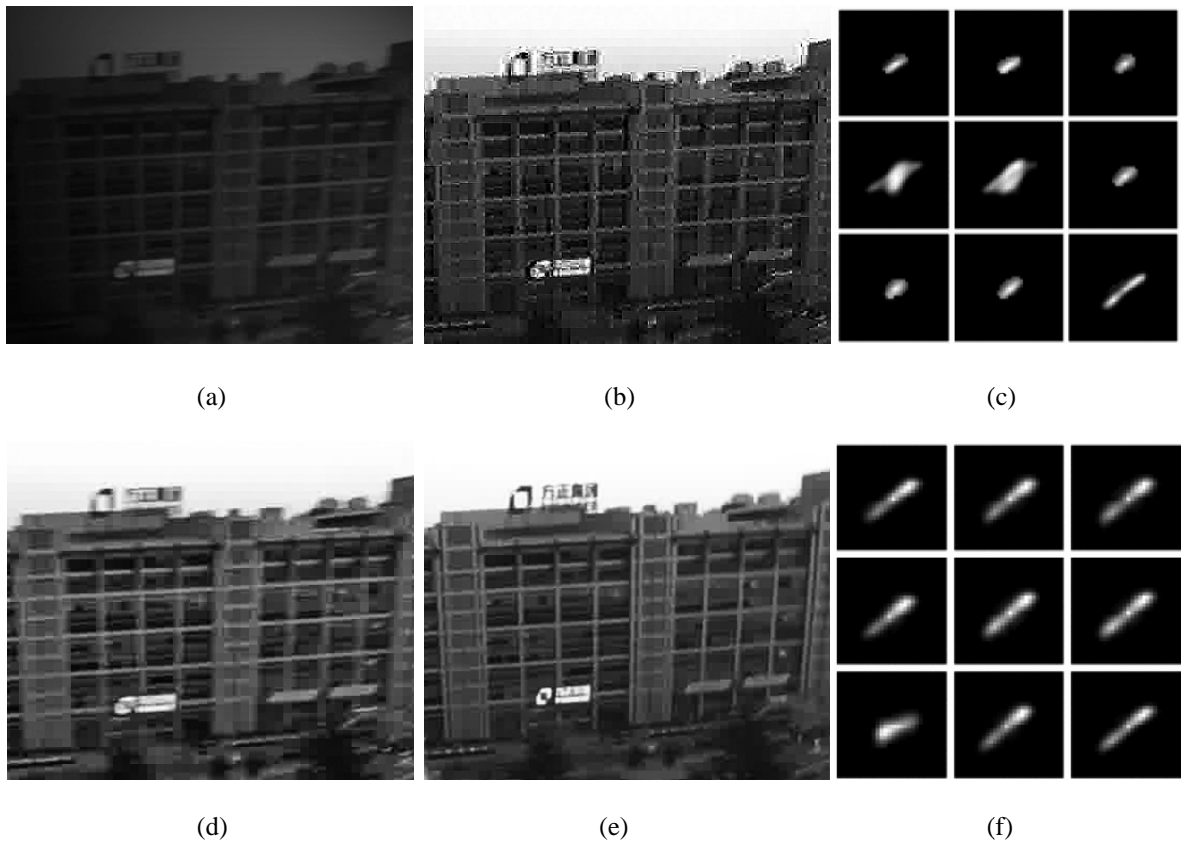

(a)                  (b)                  (c)

(d)                  (e)                  (f)

**Fig. 4** Deblurred comparison on the real data: (a) original aerial image, (b) deblurred result for (a), (c) estimated kernels for (b), (d) corrected result for (a), (e) deblurred result for (d), and (f) estimated kernels for (e).

To obtain the quantitative assessment of corrected results, the Washington DC image has been degraded by horizontal, vertical and Gaussian distributions, respectively. From the results in Table 1, the proposed method has obtained the highest values of both PSNR and SSIM. This evaluation has fully demonstrated that our non-local Retinex model can control the global intensity dispersion, preserve the details and efficiently adjust the uneven intensity distribution. In addition, the visual effect in Fig. 2 is consistent with the results of correction evaluation in Table 1. Hence it can be affirmed that our correction method outperforms other comparative methods. The effect of non-uniform correction on the deblurring quality has been evaluated quantitatively by measuring image quality. Table 2 presents the average values of PSNR and SSIM, which are both improved in the corrected and deblurred (CD) images. Because uneven lighting conditions cause the blurred image to contain more complex and low contrast makes the details more difficult to be recovered, the uncorrected and deblurred (UCD) images are relatively low-quality. In addition, the computed indices in Table 2 reflect that the blur distributions have the tendency to be uniform.

Table 1 Correction evaluation for the Washington DC image.

| Image | PSNR(dB) | | | SSIM | | |
| --- | --- | --- | --- | --- | --- | --- |
| | Horizontal | Vertical | Gaussian | Horizontal | Vertical | Gaussian |
| Mask | 32.85 | 33.13 | 33.78 | 0.955 | 0.962 | 0.966 |
| VR | 33.02 | 33.48 | 33.27 | 0.962 | 0.975 | 0.973 |
| TVR | 33.19 | 33.54 | 33.26 | 0.964 | 0.976 | 0.974 |
| Proposed | **33.56** | **33.72** | **33.94** | **0.975** | **0.981** | **0.986** |

Table 2 Quantitative evaluation for non-uniform correction effectiveness.

| Image | PSNR(dB) | | SSIM | |
| --- | --- | --- | --- | --- |
| | UCD | CD | UCD | CD |
| Washington | 31.80 | **33.01** | 0.926 | **0.958** |
| College | 32.62 | **33.97** | 0.942 | **0.981** |
| Building | 30.12 | **31.10** | 0.864 | **0.901** |

*5.3 Performance Validation and Comparison with Other Methods*

The performance of the proposed regularization scheme has been evaluated by conducting the comprehensive experiments on both synthetic and real motion blurred images. Meanwhile, our restoration method based on the Eq. (23) is qualitatively and quantitatively compared with three closely related methods: BiNorm[9], ST-NLTV[12] and SWATV[40]. Without loss of generality, we use the image set with uniform blurs to reflect the validation of the improved DNST and SA-TGV in the testing experiments. The results of all methods are obtained by the specified parameters described in the corresponding literatures. By aligning the deblurred image with the ground-truth image to compute the errors, the quality of a recovered kernel is measured by using the error ratio index.

A lake image shown in Fig. 5(a) is chosen as the groundtruth. The degraded image containing the large motion blur is obtained by using the kernel in Fig. 5(b). The white Gaussian noise with standard deviation of 0.01 is added to the blurred image. The estimated images and kernels are presented in Fig. 5 for visual comparison. It can be observed that our approach has provided the best restored result with higher sharpen degree and faithfully recovered more details in Fig. 5(f). In addition, our kernel is clean and visually accurate. Thanks to the superior structure-preserving and scale-adaptivity characteristics of the SA-TGV, the image restored by the proposed method is of high quality. On the contrary, the results of other three approaches in Fig. 5(c)-(e) have lost most useful structure information.

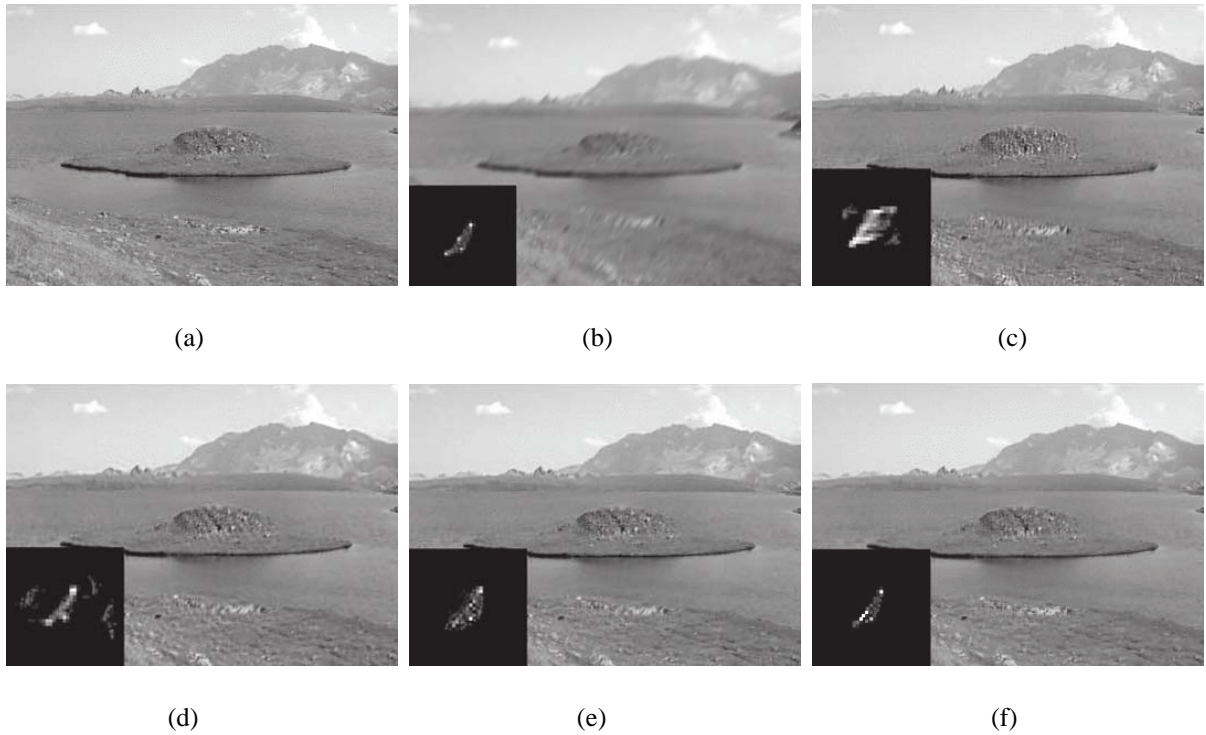

(a)                    (b)                    (c)

(d)                    (e)                    (f)

**Fig. 5** Restoration results on the synthetic image with low noise level: (a) groundtruth, (b) input degraded image, (c) BiNorm, (d) ST-NLTV, (e) SWATV, (f) our result.

The clear Aque image in Fig. 6(a) is convolved with the blur kernel given in Fig. 6(b). Then the white Gaussian noise with standard deviation of 0.05 is added to obtain the Fig. 6(b) in a controlled fashion. This experiment was conducted to explore the accuracies of restoration methods when the noise level increases and the different motions make the blur more complex.

From the comparison results shown in Fig. 6, one can see that our estimated blur kernels is closer to the ground truth, and our estimated latent image contains more natural details and less ringing artifacts than the results in Fig. 6(c)-(e). Our method is more reliably in the presence of high noise level because the improved DNST employed in our regularized terms has the powerful sensitivity and selectivity to capture and represent more geometric information of images.

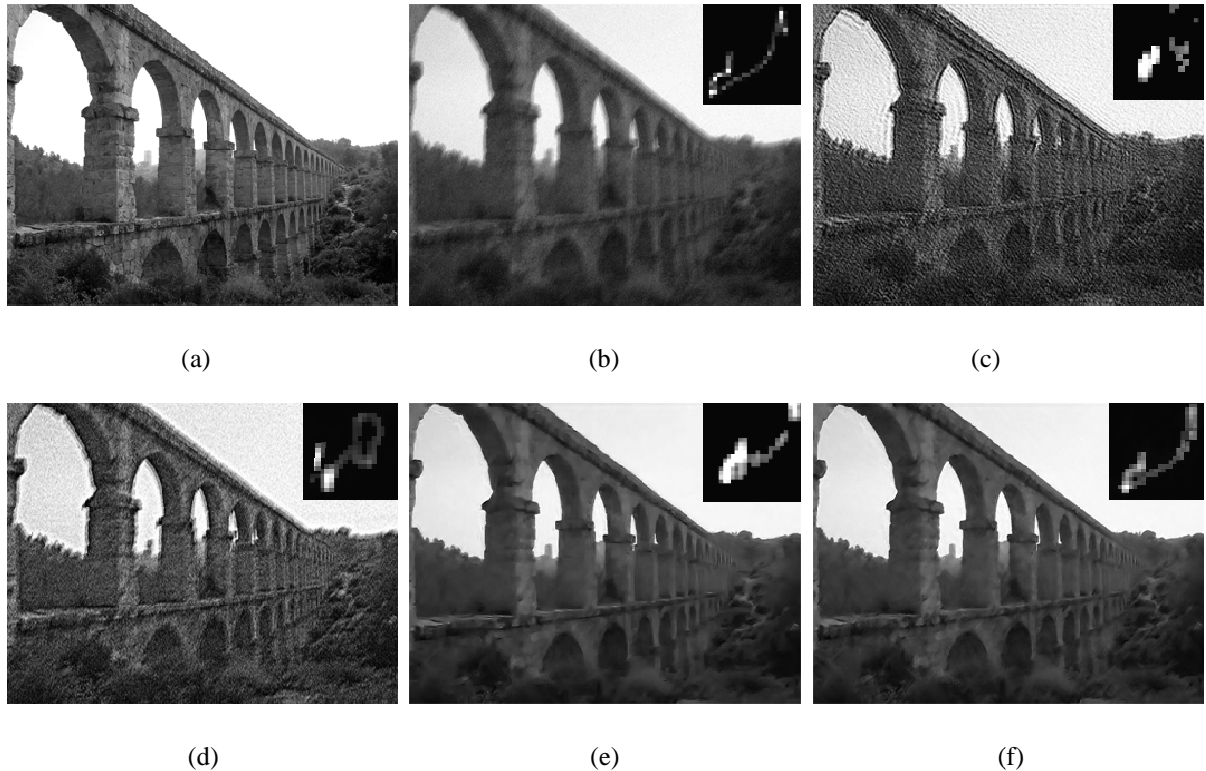

(a)                 (b)                 (c)

(d)                 (e)                 (f)

**Fig. 6** Restoration results on the synthetic image with high noise level: (a) groundtruth, (b) input degraded image, (c) BiNorm, (d) ST-NLTV, (e) SWATV, (f) our result.

The validation of our scheme has been additionally evaluated on real data set. The market image in Fig. 7(a) contains the typical blur and the noise captured in the real-world challenging condition. To manifest the visual effect, we have chosen the region cropped from Fig. 7(a) as the input image of Fig. 7(b). As shown in Fig. 7, the textures in our restored image are cleaner and sharper than those of Fig. 7(c)-(e) because the mutual contributions of the improved DNST and SA-TGV in our restoration model accommodate to represent the varying directional features in images and preserve diverse texture patterns different from the noise. In addition, our method can deduce the reliable kernel from a wide illumination range.

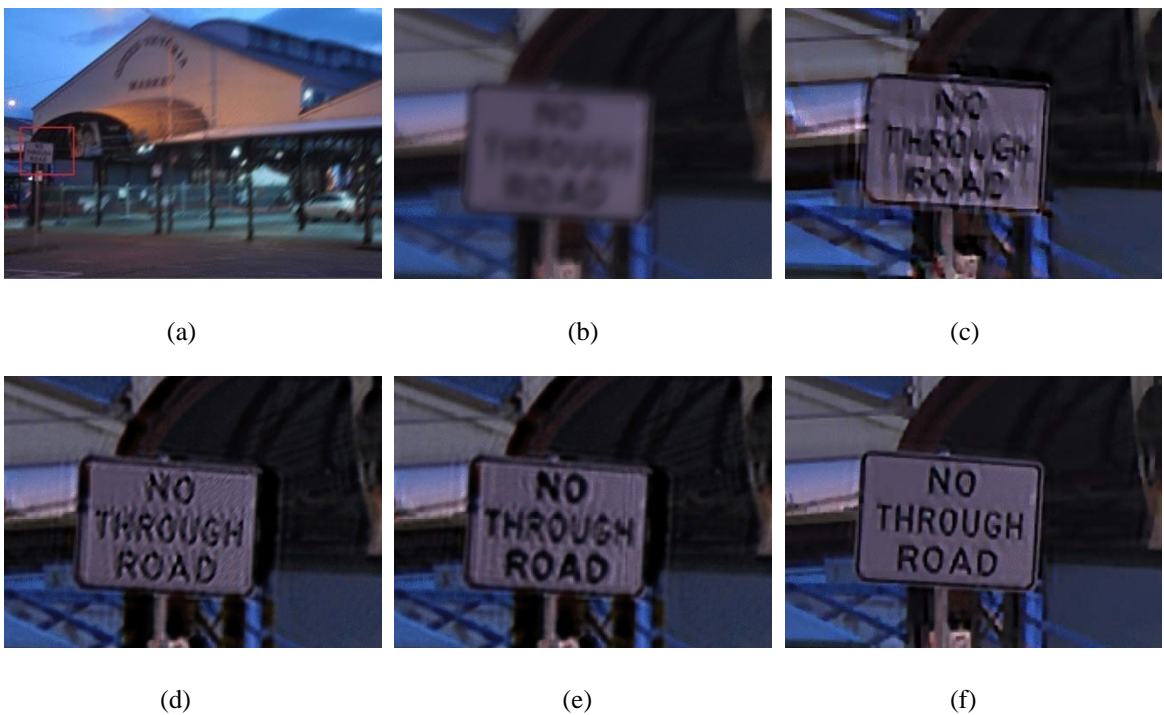

**Fig. 7** Restoration results on the real image: (a) blurry and noisy image, (b) cropped region, (c) BiNorm, (d) ST-NLTV, (e) SWATV, (f) our result.

To quantitatively measure the improvement in restored image quality, the PSNR, SSIM and error ratio are computed by using a set of images which are degraded by eight motion-blur kernels[51] for simulating the blur in aerial imaging. Also, the white Gaussian noise with the variances 1% to 10% is added to the image sequence. Table 3 shows the overall performance of the four compared methods. Our method is robust to restore the corrupted images quite well. Also it has the highest PSNR and SSIM values which are consistent to the visual improvement. The error ratios demonstrate that our kernels are more clean and accurate and the Gaussian noise could be handled well. This conclusion also can be confirmed by the results shown in Fig. 8. As can be seen, our method is more robust than other methods, and it rarely fails to recover the kernel at reasonable error ratios. In addition, the numerical computation results in the experiments show that the propose method has the fast convergence rate. And usually the exact closed-form solutions will be found within 200 iterations.

Table 3 Quantitative evaluation for the restoration performance.

|  | Average PSNR (dB) | | | Average SSIM | | | Error ratio | |
| --- | --- | --- | --- | --- | --- | --- | --- | --- |
| Image | Lake | Aque | Market | Lake | Aque | Market | Lake | Aque |
| BiNorm | 29.03 | 25.95 | 27.79 | 0.833 | 0.758 | 0.811 | 2.427 | 2.891 |
| ST-NLTV | 29.75 | 26.63 | 28.22 | 0.858 | 0.786 | 0.822 | 1.993 | 2.265 |
| SWATV | 30.81 | 28.58 | 29.31 | 0.887 | 0.827 | 0.851 | 1.516 | 1.868 |
| Proposed | **31.85** | **30.47** | **31.03** | **0.917** | **0.875** | **0.901** | **1.245** | **1.407** |

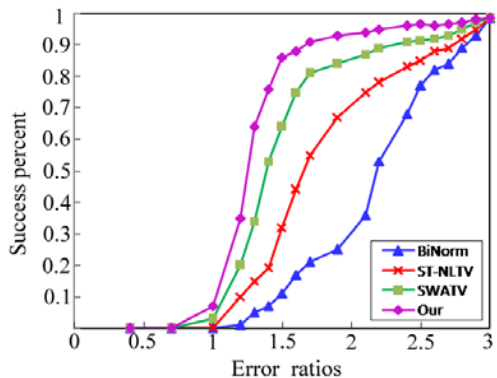

**Fig. 8** Cumulative error ratio histogram on the synthetic dataset.

## 6 Conclusion

In this paper, we propose a new blind restoration approach for aerial images via incorporating the non-local Retinex model for the non-uniform illumination correction, and the adaptive multi-scale regularization for the powerful constraints. The non-local Retinex model can correct space-variant illumination and transform non-uniform motion blur into nearly uniform kernels on certain conditions. Thus the estimated kernels achieve the smooth and sparse properties. The $\ell_1$-norms of the improved DNST coefficients and SA-TGV are jointly used as structure-dependent regularizers. This regularization strategy can render the sharpness and sparsity constraints for original image to deal with severe motion blur in varying illumination. Both the theoretical and experimental results have verified the validity and highly effectiveness of the proposed framework. Our blind restoration method can not only achieve high-quality clear aerial images but also remove the unpredictable noise and ringing artifacts, and leads to the state-of-the-art results. In the future research, our restoration method would be further extended for handling with other degraded types of aerial images.

*Acknowledgments*

Thanks to China Postdoctoral Science Foundation (2014M560852) and Major National Scientific Instrument and Equipment Development Project of China (2013YQ030967).